\documentclass{article}

\usepackage{times}
\usepackage{graphicx} % more modern
\usepackage{subfigure}

\usepackage{natbib}

\usepackage{algorithm}
\usepackage{algorithmic}

\usepackage{amsmath}

\usepackage{multirow}
\usepackage{multicol}

\usepackage{hyperref}

\usepackage{float}

\usepackage[accepted]{icml2016}

\icmltitlerunning{Deep Structured Energy Based Models for Anomaly Detection}

\begin{document}

\twocolumn[ \icmltitle{Deep Structured Energy Based Models for Anomaly Detection}

% unless you've provided the [accepted] option to the icml2013
\icmlauthor{Shuangfei Zhai$^{*\dag}$}{szhai2@binghamton.edu}
\icmlauthor{Yu Cheng$^{*\S}$}{chengyu@us.ibm.com}
\icmlauthor{Weining Lu$^{\ddag}$}{luwn14@mails.tsinghua.edu.cn}
\icmlauthor{Zhongfei (Mark) Zhang$^{\dag}$}{zhongfei@cs.binghamton.edu }
\icmladdress{$^{\dag}$Binghamton University, Vestal, NY 13902, USA. \\ $^{\S}$IBM T. J. Watson Research Center, Yorktown Heights, NY 10598, USA. \\ $^{\ddag}$Tsinghua University, Beijing 10084, China.
\\ $^{*}$Equal contribution}

\icmlkeywords{EBM, anomaly detection}

\vskip 0.3in ]

\begin{abstract}
In this paper, we attack the anomaly detection problem by directly modeling the data distribution with deep architectures. We propose deep structured energy based models (DSEBMs), where the energy function is the output of a deterministic deep neural network with structure. We develop novel model architectures to integrate EBMs with different types of data such as static data, sequential data, and spatial data, and apply appropriate model architectures to adapt to the data structure. Our training algorithm is built upon the recent development of score matching \cite{sm}, which connects an EBM with a regularized autoencoder, eliminating the need for complicated sampling method. Statistically sound decision criterion can be derived for anomaly detection purpose from the perspective of the energy landscape of the data distribution. We investigate two decision criteria for performing anomaly detection: the energy score and the reconstruction error. Extensive empirical studies on benchmark tasks demonstrate that our proposed model consistently matches or outperforms all the competing methods.
\end{abstract}

\section{Introduction}
\label{sec:intro}
% * <zhaisf@gmail.com> 2015-12-29T18:24:36.252Z:
%
% introduction to anomaly detection
%
% ^ <zhaisf@gmail.com> 2015-12-29T21:45:04.211Z.
Anomaly detection (also called novelty or outlier detection) is to identify patterns that do not conform to the expected normal patterns \cite{survey}. Existing methods for outlier detection either construct a profile for normal data examples and then identify the examples not conforming to the normal profile as outliers, or explicitly isolate outliers based on statistical or geometric measures of abnormality. A variety of methods can be found in the survey \cite{Zimek:2012:SUO:2388912.2388917}. Anomaly detection is to correctly characterize data distribution in nature, so the normality of the data characteristic can be characterized as a distribution and any future data can be benchmarked against the normality. Apparently, the statistical power and accuracy of the anomaly detection methods depend on the capacity of the model that is used to characterize the data distribution. 

Our work is inspired by the extraordinary capacity of deep models which are able to capture the complex distributions in real-world applications. Recent empirical and theoretical studies indicate that deep architectures are able to achieve better generalization ability compared to the shallow counterparts on challenging recognition tasks \cite{Bengio2009}. The key ingredient to the success of deep learning is its ability to learn multiple levels of representations with increasing abstraction. For example, it is shown that properly regularized autoencoders \cite{dae,cae} are able to effectively characterize the data distribution and learn useful representations, which are not achieved by shallow methods such as PCA or K-Means.

However, deep models have not been systematically studied and developed for anomaly detection. Central questions critical to this development include: 1) how to effectively model the data generating distribution with a deep model? 2) how to generalize a model to a range of data structures such as static data, sequential data, and spatial data? 3) how to develop computationally efficient training algorithms to make them scalable? and 4) how to derive statistically sound decision criteria for anomaly detection purpose? 

To answer these questions in a systematic manner, in this work, we propose deep structured energy based models (DSEBMs). Our approach falls into the category of energy based models (EMBs) \cite{ebm}, which is a powerful tool for density estimation. An EBM works by coming up with a specific parameterization of the negative log probability, which is called energy, and then computing the density with a proper normalization. In this work, we focus on deep energy based models \cite{debm}, where the energy function is composed of a deep neural network. Moreover, we investigate various model architectures as to accommodate different data structures. For example, for data with static vector inputs, standard feed forward neural networks can be applied. However, for sequential data, such as audio sequence, recurrent neural networks (RNNs) are known to be better choices. Likewise, convolutional neural networks (CNNs) are significantly more efficient at modeling spatial structures \cite{alexnet}, such as on images. Our model thus allows the energy function to be composed of deep neural networks with designated structures (fully connected, recurrent or convolutional), significantly extending the application of EBMs to a wide spectrum of data structures.

Despite its powerful expressive ability, the training of EBMs remains fairly complicated compared to the training of a deterministic deep neural network, as the former requires carefully designed sampling algorithm to deal with its intrinsic intractability. The need of efficient training algorithms is even more severe with DSEBMs, given its complicated structured parameterization. To this end, we adopt the score matching method \cite{sm} as the training algorithm, instead of the default maximum likelihood estimation (MLE). Similarly to \cite{smae1}, we are able to train a DSEBM in the same way as that of a deep denoising autoencoder (DAE) \citet{dae}, which only requires standard stochastic gradient descent (SGD). This significantly simplifies the training procedure and allows us to efficiently train a DSEBM on large datasets. 

In order to perform the actual anomaly detection with a trained DSEBM, we investigate two decision criteria, the energy score and the reconstruction error. We show that the two criteria are closely connected from the view of the energy landscape, and evaluate their effectiveness under different scenarios. We perform extensive evaluations on several benchmarks varying from static, sequential to spatial data. We show that our method consistently matches or outperforms the competing algorithms.

\section{Background}
\subsection{Energy Based Models (EBMs)}
EBMs are a family of probabilistic models that can be used to build
probability density functions. An EBM parameterizes a density
function for input $x\in R^d$ in the form:
\begin{equation}
\label{eq:ebm} p(x;\theta) = \frac{e^{-E(x;\theta)}}{Z(\theta)},
\end{equation}
where $E(x;\theta)$ is the energy (negative log probability) associated with instance $x$;
$Z(\theta) = \int_x{e^{-E(x;\theta)}}dx$ is the partition function
to ensure that the density function integrates to probability $1$;
$\theta$ are the model parameters to be learned. The nice property of
EBM is that one is free to parameterize the energy in any sensible
way, giving it much flexibility and expressive power. Learning
is conducted by assigning lower energy (hence higher probability) to
observed instances and vice versa. However, directly applying MLE is impossible
due to the intractability of the partition fuction $Z(\theta)$, and thus one usually needs to resort to MCMC methods and approximate the integration with the
summation over samples from a Markov chain.

\subsection{Restricted Boltzmann Machine (RBM)}
RBM \cite{rbm} is one of the most well know examples of EBM. For
continuous input data, the energy function of an RBM takes the form:
\begin{equation}
\label{eq:rbm} 
E(x;\theta) = \frac{1}{2}\|x - b'\|_2^2 -
\sum_{j=1}^K g(W_j^T x + b_j),
\end{equation}
where $W \in R^{d \times K}$ (with $W_j$ being the $j$th column), $b \in R^K$ (with $b_j$ as the $j$th element) and $b' \in R^d$ are the parameters to learn; $g(x)$ is the soft plus function $\log(1 + e^x)$.
Multiple RBMs can be trained and stacked on top of each other to formulate a deep RBM,
which makes it useful for initializing multi-layer neural networks.
Although efficient training algorithms, such as contrastive
divergence, are proposed to make RBM scalable, it is still
considerably more difficult to train than a deterministic
neural network \cite{rbm}.

\subsection{Denoising Autoencoders and Score Matching}
Autoencoders are unsupervised models that learn to reconstruct the
input. A typical form of autoencoder minimizes the following objective function:
\begin{equation}
\label{eq:ae}
\begin{split}
\sum_{i=1}^N \|x_i - f(x_i; \theta)\|_2^2
\end{split}
\end{equation}
where $f(\cdot; \theta)$ is the reconstruction function that maps $R^d \rightarrow
R^d$ which is usually composed of
an encoder followed by a decoder with symmetrical architecture and shared parameters. One particularly interesting variant of
autoencoders is DAEs \cite{dae}, which learn to construct the inputs
given their randomly corrupted versions:
\begin{equation}
\label{eq:dae} \sum_{i=1}^N \mathrm{E}_{\epsilon}\|x_i - f(x_i +
\epsilon; \theta)\|_2^2,
\end{equation}
where $\epsilon \sim \mathcal{N}(0, \sigma^2I)$ is an isotropic
Gaussian noise. DAEs are easy to train with standard
stochastic gradient descent (SGD) and perform significantly better
than unregularized autoencoders.

While RBM and DAEs are typically considered as two alternative unsupervised deep models, it is recently shown they are closely related to each other. In particular, \cite{smae1} shows that training an RBM with score
matching (SM) \cite{sm} is equivalent to a one-layer DAE. SM is an
alternative method to MLE, which is especially suitable for
estimating non-normalized density functions such as EBM. Instead of
trying to directly maximize the probability of training instances,
SM minimizes the following objective function:
\begin{equation}
\label{eq:sm} J(\theta) = \frac{1}{2}\int_x p_x(x)\|\psi(x;\theta) -
\psi_x(x)\|^2_2dx,
\end{equation}
where $p_x(x)$ is the true data distribution which is unknown;
$\psi(x;\theta) = \nabla_{x}\log p(x;\theta) = -\nabla_x
E(x;\theta)$ and $\psi_x(x) = \nabla_{x}\log p_x(x)$ are the
score function of the model and the true density function, respectively.
In words, $J(\theta)$ measures the expected distance of the scores
between the model density and the true density. \cite{smae1} shows
that by approximating the $p_x(x)$ with the Parzen window density $\frac{1}{N}\sum_{i=1}^N \mathcal{N}(x_i, \sigma^2 \mathbf{I})$, minimizing Equation \ref{eq:sm} yields an objective
function in the same form as that of an autoencoder in Equation \ref{eq:dae}, with a reconstruction function defined as:
\begin{equation}
\label{eq:f}
\begin{split}
&f(x;\theta) = x - \nabla_{x}E(x;\theta).
\end{split}
\end{equation}
Substituting $E(x;\theta)$ with Equation \ref{eq:rbm} yields a typical one-layer DAE (up to a constant factor):
\begin{equation}
\begin{split}
f(x;\theta) = W \sigma(W^T x + b) + b',
\end{split}
\end{equation}
where $\sigma(x)$ is the sigmoid function $\frac{1}{1 + e^{-x}}$.

Equation \ref{eq:f} plays a key role in this work, as it allows one to efficiently train an arbitrary EBM in a similar way to that of a DAE, as long as the energy function is differentiable w.r.t. $x$ and $\theta$. As will be demonstrated below, this makes training a deep EBM with various underlying structures in an end-to-end fashion, without the need of resorting to sophisticated sampling procedures \cite{debm} or layer wise pretraining \cite{rbm}.

\section{Deep Structured EBMs}
Deep architectures allow one to model complicated patterns efficiently, which makes it especially suitable for high dimensional data. On the other hand, it is often necessary to adapt the architecture of a deep model to the structure of data. For example, recurrent neural networks (RNNs) have been shown to work very well at modeling sequential data; convolutional neural networks (CNNs) are very effective at modeling data with spatial structure. In this work, we extend EBMs further to deep structured EBMs (DSEBMs), where we allow the underlying deep neural network encoding the energy function to take architectures varying from fully connected, recurrent, and convolutional. This generalizes EBMs \cite{ebm} as well as deep EBMs \cite{debm} as it makes our model applicable to a much wider spectrum of applications, including static data, sequential data and spatial data. Moreover, we show that by deriving the proper reconstruction functions with Equation \ref{eq:f}, DSEBMs can be easily trained with SGD, regardless of the type of the underlying architecture. In the following, we will elaborate the formulation of DSEBMs for the three cases.
\subsection{Fully Connected EBMs}
This case is conceptually the same as the deep EBMs proposed in \cite{debm}. Without loss of generality, we express the energy function of an L-layer fully connected EBM as:
\begin{equation}
\label{eq:deepebm}
\begin{split}
&E(x;\theta) = \frac{1}{2}\|x - b'\|_2^2 - \sum_{j=1}^{K_L} h_{L,j} \\
s.t. \; &h_l = g(W_l^T h_{l-1} + b_1), \; l \in [1, L]
\end{split}
\end{equation}
where $W_l \in R^{K_{l-1} \times K_{l}}, b_l \in R^{K_{l}}$ are the parameters for the $l$th layer; $K_{l}$ is the dimensionality of the $l$th layer. The $0$th layer is defined as the input itself; and thus we have $K_0 = d, h_0 = x$. We have explicitly included the term $\|x - b'\|_2^2$ which acts as a prior, punishing the probability of the inputs that are far away from $b' \in R^d$. 

Following the chain rule of gradient computation, one can derive the reconstruction function as follows:
\begin{equation}
\label{eq:deepdae}
\begin{split}
&f(x;\theta) = x - \nabla_x{E(x;\theta)} = h_0' + b'\\
&h'_{l-1} = \frac{\partial h_l}{\partial h_{l-1}} h'_{l} =  \sigma(W_l^T h_{l-1} + b_l) \cdot (W_l h'_{l}), \\
& for \; l \in [1, L-1],
\end{split}
\end{equation}
where $h'_{L} = \mathbf{1}$ with $\mathbf{1} \in R^{K_L}$ denoting a column vector of all ones; $\cdot$ is the element-wise product between vectors. One can then plug in the resulting $f(x;\theta)$ into Equation \ref{eq:dae} and train it as a regular L-layer DAE.

\subsection{Recurrent EBMs}
Our formulation of recurrent EBMs is similar to \cite{rnnrbm}, where an EBM is built at each time step, with parameters determined by an underlying RNN. Formally, given a sequence of length $T$
$\mathbf{x} = [x^1, ..., x^T], x^t \in R^d$, we factorize the joint probability as
$p(\mathbf{x}) = \prod_{t=1}^T p(x^t|x^{1, ..., t-1})$ with the chain rule of probability. For
each time step $t$, $p(x^t|x^{1, ..., t-1})$ is modeled as
an EBM with energy $E(x^t|\theta^t)$. In contrast to the conventional formulation of EBM, the parameters of the EMB $\theta^t$ is a function of the inputs from all the previous time steps $x^{1, ..., t-1}$. 
A natural choice of $\theta^t$ is to let it be the output of an RNN. As a concrete example, consider the energy function at each step follows that in Equation \ref{eq:rbm}, by replacing $W, b, b'$ with $W^t, b^t, b'^t$. Directly letting the RNN to update all the parameters at each step requires a large RNN with lots of parameters. As a remedy, \cite{rnnrbm} proposes to fix $W^t=W$ for all time steps, only letting $b^t$ and $b'^t$ to be updated by the RNN:
\begin{equation}
\label{eq:rnnrbm}
\begin{split}
& h^t = g(W_{hh}h^{t-1} + W_{hx}x^t + b_h) \\
&b^t = W_{bh}h^t + b, \; b'^t = W_{b'h} h^t + b',
\end{split}
\end{equation}
where $W_{hh} \in R^{K_{rnn} \times K_{rnn}}, W_{hx} \in R^{K_{rnn} \times d}, b_h \in R^{K_{rnn}}$ are parameters of the RNN; $W_{bh} \in R^{K_{ebm} \times K_{rnn}}, b \in R^{K_{ebm}}, W_{b'h} \in R^{d \times K_{rnn}}, b' \in R^d$ are the weights with which to transform the hidden state of the RNN to the adaptive biases.

Training the recurrent EBM with score matching is similar to that of a fully connected EBM. To see this, we now have $p(\mathbf{x}) = \frac{e^{-\sum_{t=1}^TE(x^t;\theta^t)}}{\prod_{t=1}^TZ^t}$, where $Z^t = \sum_x e^{-E(x;\theta^t)}$ is the partition function for the $t$th step. Plugging $p(\mathbf{x})$ into Equation \ref{eq:sm}, we have $\psi(\mathbf{x}) = \nabla_{\mathbf{x}} \log p(\mathbf{x}) \approx -[\nabla_{x^1}^TE(x^1;\theta^1), ..., \nabla_{x^T}^TE(x^T;\theta^T)]^T$. In the last step, we have made a simplification by omitting the gradient term of $\nabla_{x^j} E(x^i;\theta^i)$, for $j < i$ \footnote{While one can also choose to use the full gradient, we find this simplification works well in practice, yielding an objective in a much more succinct form.}. Accordingly, Equation \ref{eq:f} is modified to $f(\mathbf{x}) = \mathbf{x}  -[\nabla_{x^1}^TE(x^1;\theta^1), ..., \nabla_{x^T}^TE(x^T;\theta^T)]^T$. One is then able to train the recurrent EBM by plugging $f(\mathbf{x})$ into Equation \ref{eq:dae} and perform the standard SGD. Note that in this case the standard backpropagation is replaced with backpropagation through time, due to the need of updating the RNN parameters. 

\subsection{Convolutional EBMs}
Previously, the combination of CNN and RBM have been proposed in
\cite{cnnrbm}, where several layers of RBMs are alternately convolved with an image then stacked on top of each other. In this paper, we take a significantly different approach by directly building deep EBMs with convolution operators (with optional pooling layers or fully connected layers), simply by replacing $h_L$ in Equation \ref{eq:deepebm} with the output of a CNN. Using a deterministic deep convolutional EBM allows one to directly train the model end-to-end with score matching, thus significantly simplifies the training procedure compared with \cite{cnnrbm}. Formally, consider the input of the $(l - 1)$th layer $h_{l-1} \in R^{K_{l-1} \times d_{l-1} \times d_{l-1}}$  is a $d_{l-1} \times d_{l-1}$ image with $K_{l-1}$ channels. We define $h_{l}$ as the output of a convolution layer:
\begin{equation}
\label{eq:cnnrbm}
\begin{split}
h_{l, j} = g(\sum_{k=1}^{K_{l-1}}\tilde{W}_{l, j, k} \ast h_{l-1, k} + b_{l, j}), \; j \in [1, K_l].
\end{split}
\end{equation}
Here $W \in R^{K_l  \times K_{l-1} \times d_{w,l} \times d_{w,l}}$ are the $K_l$ convolutional filters of size $d_{w,l}\times d_{w,l}$; $b_l \in R^{K_l}$ is the bias for each filter. We denote the
tilde operator ($\tilde{A}$) as flipping a matrix $A$ horizontally
and vertically; $\ast$ is the "valid" convolution operator, where convolution is only conducted where the input and the filter
fully overlap. $d_l = d_{l-1} - d_{w,l} + 1$ is thus the size of the
output image following the valid convolution. In order to compute the reconstruction function following equation \ref{eq:deepdae}, we modify the recurrence equation from $h'_l$ to $h'_{l-1}$ for a convolutional operator as: 
\begin{equation}
\label{eq:cnngrad}
h'_{l-1, k} = \sum_{j=1}^{K_l}[\sigma(\sum_{k=1}^{K_{l-1}}\tilde{W}_{l, j, k} \ast h_{l-1, k} + b_{l, j})\cdot (  W_{l,j, k} \odot h'_{l, j})].
\end{equation}
Here we have denoted $\odot$ as the "full" convolution operator,
where convolution is conducted whenever the input and the filter
overlap by at least one position. Besides the convolution layer, we can also use a max pooling layer which typically follows a convolution layer. Denote $d_p \times d_p$ as the pooling window size, passing $h_{l-1}$ through a max pooling layer gives an output as:
\begin{equation}
\label{eq:maxpooling}
\begin{split}
&h_{l,k,p,q},\; x_{l,k,p,q}, y_{l, k, p,q} \\ 
&= \max_{(p-1)d_p + 1 \leq i \leq pd_p, (q-1)d_p + 1 \leq j \leq qd_p}{h_{l-1,k, i, j}}.
\end{split}
\end{equation}
Here the $\max$ operation returns the maximum value as the first term and the corresponding maximum coordinates as the second and third. Rewriting the recurrence equation corresponding to Equation \ref{eq:deepebm} yields:
\begin{equation}
\label{eq:maxgrad}
h'_{l-1,k, x_{l,k,p,q}, y_{l,k,p,q}} = h'_{l, k, p, q}, \; p,q \in [1, d_l],
\end{equation}
where the unassigned entries of $h'_{l-1}$ are all set as zero.

The derivation above shows that one is able to compute the reconstruction function $f(x)$ for a deep convolutional EBM consisting of convolution, max pooling and fully connected layers. Other types of layers such as mean pooling can also be managed in a similar way, which is omitted due to the space limit. 

\section{Deep Structured EBMs for Anomaly Detection}

Performing anomaly detection given a trained EBM naturally corresponds to identifying data points that are assigned low probability by the model. With a trained DSEBM, we can then select samples that are assigned
probability lower than some pre-chosen threshold $p_{th}$ as outliers. Although computing the exact probability according to Equation
\ref{eq:ebm} is intractable, one can immediately recognize the following logic:
\begin{equation}
\label{eq:th_energy}
\begin{split}
&p(x;\theta) < p_{th} \Rightarrow \log p(x;\theta) < \log p_{th}
\Rightarrow \\ &E(x; \theta) > \log p_{th} + \log Z(\theta)
\Rightarrow E(x; \theta) > E_{th}.
\end{split}
\end{equation}
Here we have used the fact that $Z(\theta)$ is a constant that does
not depend on $x$; hence selecting samples with probability lower than $p_{th}$ is equivalent to selecting those with energy higher than a corresponding energy threshold $E_{th}$. \footnote{For a recurrent DSEBM, this decision rule is modified as $\sum_{t=1}^T E(x^t; \theta^t) > E_{th}$} 

\begin{figure}[t]
\includegraphics[width=0.45\textwidth,height=0.36\textwidth]{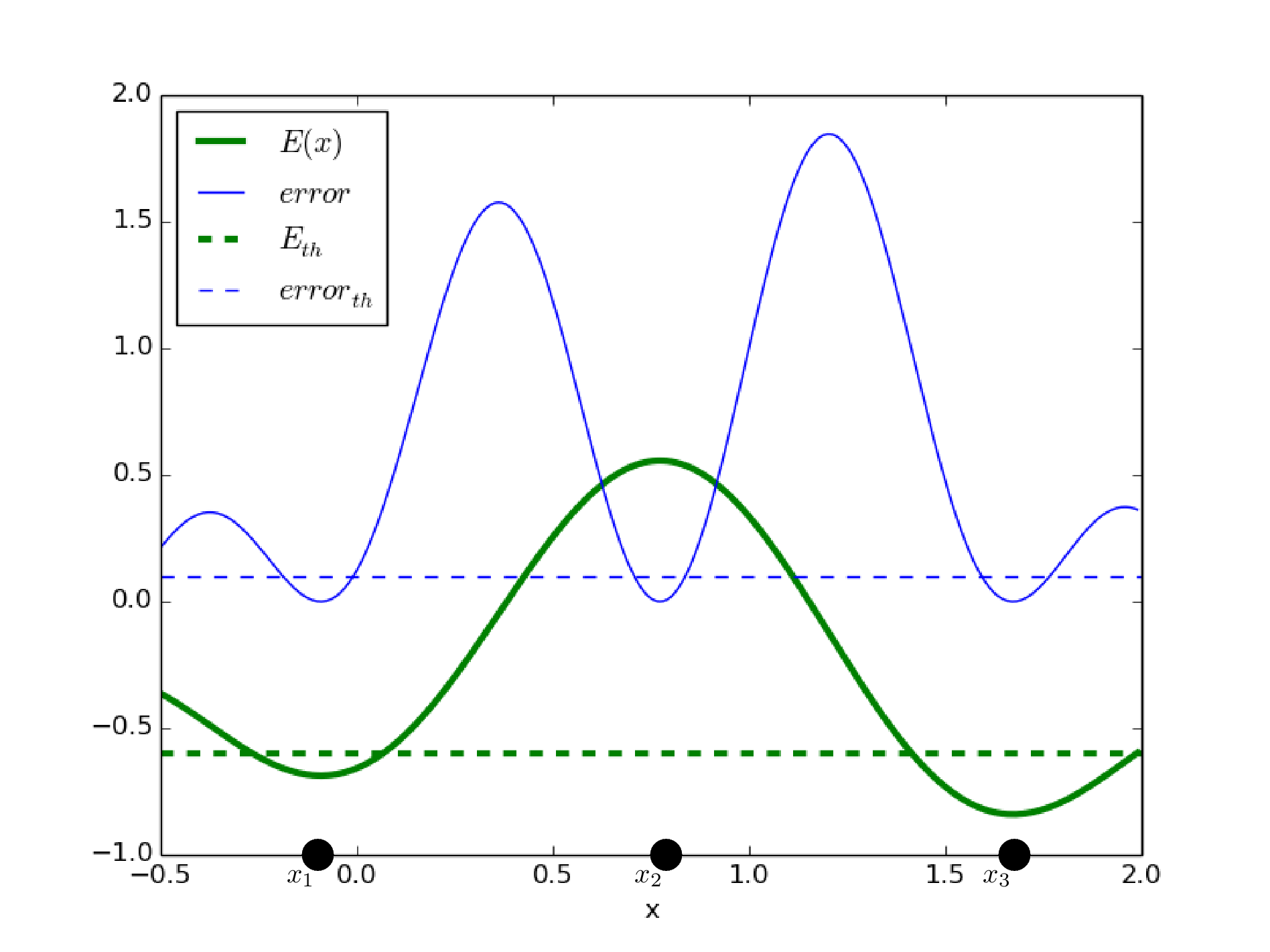}
\label{fig:energy}
\caption{A 1D demonstration of using energy $E(x)$ and the reconstruction error $\|\nabla_x E(x;\theta)\|_2^2$ (denoted as $error$ in the figure) as the decision criterion. For each of the two criteria, samples with value (energy or reconstruction error) above the chosen threshold are identified as outliers.}
\end{figure}

Moreover, motivated by the connection of EBMs and DAEs, we further investigate another decision criteria which is based on the reconstruction error. In the early work of \cite{conf/icdm/WilliamsBHHG02}, autoencoders (which they call replicator neural networks) have been applied to anomaly detection. However, their adoption autoencoders are unregularized, and thus could not be interpreted as a density model. The authors then propose to use the reconstruction error as the decision criterion, selecting those with high reconstruction error as outliers. On the other hand, with a DSEBM trained with score matching, we are able to derive the corresponding reconstruction error as $\|x - f(x;\theta)\|^2_2 = \|\nabla_x E(x;\theta)\|^2_2$. The corresponding decision rule is thus $\|\nabla_x E(x;\theta)\|^2_2 > Error_{th}$, with some threshold $Error_{th}$. \footnote{For a recurrent DSEBM, this decision rule is modified as $\sum_{t-1}^T \|\nabla_{x^t}E(x^t; \theta^t)\|^2_2 > Error_{th}$} In other words, examples with high reconstruction errors correspond to examples whose energy has large gradient norms. This view makes reconstruction error a sensible criterion as on the energy surface inliers usually sit close to local minimums, where the gradient should be close to zero. However, using the reconstruction error might produce false positive examples (outliers that are classified as inliers), as if a sample sits close to a local maximum on the energy surface, the gradient of its energy will also be small. We demonstrate the two criteria in Figure \ref{fig:energy} with a 1D example. As we see, for $x_1$ and $x_3$, both energy and reconstruction error produces the correct prediction. However, $x_2$ is a false positive example under the reconstruction error criterion, which energy correctly recognizes as an outlier. However, note that in a high dimensional input space, the probability that an outlier resides around a local maximum grows exponentially small w.r.t. the dimensionality. As a result, the reconstruction error still serves as a reasonable criterion.   

\begin{table*}[t]
\centering 
\small
\begin{tabular}{|c | c c c c c|}
\hline & \# Classes &  \# Dimensions &  \# Instances & \rm{Avg}(t) & Outlier ratio ($\rho$) \\
\hline KDD99 \cite{Lichman:2013}  & 2 & 41 & 494,021 & \rm{NA} & $\rho=0.2$\\
Thyroid   & 3 & 10  & 7,200 &  \rm{NA} & $\rho=0.1$ \\
Usenet  & 2 & 659 & 5,931 &  \rm{NA} & $\rho=0.5$ \\
\hline 
CUAVE \cite{Patterson02cuave:a}  & 10 & 112 & 1,790 & 45.86 & $0.1 \le \rho \le 0.4$\\
NATOPS \cite{Patterson02cuave:a} & 20 & 20 & 9,600 & 49.51 & $0.1 \le \rho \le 0.4$\\
FITNESS & 2 & 20 & 6,031 & 405.39 & $0.1 \le \rho \le 0.4$\\
\hline
Caltech-101 \cite{Fei-Fei:2007:LGV:1235884.1235969} & 101 & 300 $\times$ 200  & 9,146 & NA & $0.1 \le \rho \le 0.4$ \\
MNIST \cite{Lecun98gradient-basedlearning}  & 10 & 28 $\times$ 28 & 70,000 & NA & $0.1 \le \rho \le 0.4$ \\ 
CIFAR-10 \cite{Krizhevsky09}  & 10 & 32 $\times$ 32 & 60,000 & NA & $0.1 \le \rho \le 0.4$ \\
\hline
\end{tabular}
\caption{Specification of benchmark data sets we used. \rm{Avg}(t) is the length of sequence for sequential data.} \label{table:benchmark}
\end{table*}

\section{Experimental Evaluation}
In this section, we evaluate the proposed anomaly detection framework, where our two proposed anomaly detection criteria using energy and reconstruction error are abbreviated as DSEBM-e and DSEBM-r, respectively. Our experiments consist of three types of data: static data, sequential data (e.g., audio) and spatial data (e.g., image), where we apply fully connected EBM, recurrent EBM and convolutional EBM, respectively. The specifications of benchmark datasets used are summarized in Table \ref{table:benchmark}. To demonstrate the effectiveness of DSEBM, we compare our approach with several well-established baseline methods that are publicly available. Ground truth labels are available in all data sets; and we report precision (mean precision), recall (mean recall), and $F_{1}$ score (mean $F_{1}$) for the anomaly detection results achieved by all methods. Below we detail the baselines, explain the experimental methodology, and discuss the results.

\begin{table*}
\small
\centering
\begin{tabular}{|c|ccc|ccc|ccc|}
\hline
\multirow{2}{*}{\textbf{Method}} & \multicolumn{3}{c|}{KDD99} & \multicolumn{3}{c|}{Thyroid}  & \multicolumn{3}{c|}{Usenet} \\
\cline{2-10}
& \textbf{Presion} & \textbf{Recall} & \textbf{$F_1$} & \textbf{Presion} & \textbf{Recall} & \textbf{$F_1$} & \textbf{Presion} & \textbf{Recall} & \textbf{$F_1$} \\
\hline
PCA & 0.8312 & 0.6266 & 0.7093 & 0.9258 &0.7322 & 0.8089 & 0.7025 & 0.7648 & 0.7211 \\ 
Kernel PCA & \bf 0.8627 & 0.6319 & 0.7352 & 0.9537 & 0.7493 & 0.8402 & 0.7215 & 0.7930 & 0.7387 \\
KDE & 0.8119 & 0.6133 & 0.6975 & 0.9275 & 0.7129 & 0.7881 & 0.6749 & 0.7535 & 0.7080 \\
RKDE  & 0.8596 & 0.6328 & 0.7322 & 0.9437 & 0.7538 & 0.8429 & \bf 0.7248 & 0.7855 & 0.7402 \\
OC-SVM  & 0.8050 & \bf 0.6512 & 0.7113 & \bf 0.9602 & 0.7424 & \bf 0.8481  & 0.6978 & 0.7795 & 0.7320 \\
AEOD & 0.7624 & 0.6218 & 0.6885 & 0.9157 & 0.6927 & 0.7873 & 0.6095 & 0.7579 & 0.6741 \\
DSEBM-r & 0.8521 &  0.6472 & 0.7328 & 0.9527 & 0.7479 & 0.8386 & 0.7205 & 0.7837 & 0.7314 \\
DSEBM-e & 0.8619 & 0.6446 & \bf 0.7399 & 0.9558 & \bf 0.7642 & 0.8375 & 0.7129 & \bf 0.8081 & \bf 0.7475 \\
\hline
\end{tabular}
\caption{\textbf{KDD99, Thyroid, Usenet}: precision, recall and $F_1$ over over the static data sets of seven methods. For each column, the best result is shown in boldface.}
\label{tab:result:static}
\end{table*}

\subsection{Static Data}
There benchmark datasets are used in this study: KDD99 10 percent, Thyroid and Usenet from the UCI repository \cite{Lichman:2013}. The training and test sets are split by 1:1 and only normal samples are used for training the model. We compare DSEBMs (with 2-layer fully connected energy function) with a variety of competing methods, including two reconstruction-based outlier detection methods, PCA and Kernel PCA, two density-based methods Kernel Density Estimator (KDE) \cite{parzen1962estimation} and Robust Kernel Density Estimator (RKDE) \cite{DBLP:journals/jmlr/KimS12}, along with the traditional one-class learning method One-Class SVM (OC-SVM) \cite{Scholkopf:2001:ESH:1119748.1119749}. We also include the method proposed in \cite{conf/icdm/WilliamsBHHG02}, named AutoEncoder Outlier Detection (AEOD) as one baseline. The results are shown in Table \ref{tab:result:static}. We see that, overall, DSEBM-e and DSEBM-r achieve comparable or better performances compared with the best baselines. On Thyroid, the performances of DSEBMs are slightly worse than OC-SVM and RKDE. We speculate that this is most likely caused by the low dimensionality of the dataset (10), where kernel based methods (which OC-SVM and RKDE are) are very effective. However, on Usenet which has a much higher dimensionality (659), DSEBM-e achieves the best result, measured by recall and $F_1$. This is consistent with our intuition, as on high-dimensional datasets, deep models are more effective and necessary to resolve the underlying complication. On KDD99, DSEBM-e also achieves the best $F_1$.

\subsection{Sequential Data}
For this task, we use three sequential datasets: (1) CUAVE which contains audio-visual data of ten spoken digits (zero to nine); (2) NATOPS which contains 24 classes of body-and-hand gestures used by the US Navy in aircraft handling aboard aircraft carriers; (3) FITNESS which contains users' daily fitness behaviors collected from health care devices, including diet, sleep and exercise information. According to the BMI change, the users are categorized into two groups "losing weight" and "gaining weight". For a single category, the outlier samples are simulated with a proportion $0.1 \le \rho \le 0.4$ from other categories. The datasets are split into training and test by 2:1, where $2/3$ of the normal samples are used for training split. We compare DSEBM-r and DSEBM-e with three static baselines, Kernel PCA, RKDE and OC-SVM. Also, we include two sequential methods: 1) HMMs, where the model is trained with the normal training sequences, and the posterior probability $p(y|x)$ of each test sequence is computed as the normalized negative log-likelihood; 2) OCCRF \cite{Song:2013:OCR:2540128.2540370}, where the model learns from a one-class dataset and captures the temporal dependence structure using conditional random fields (CRFs). Table \ref{tab:sequential:result} (with $\rho=0.3$) shows the performances of all the methods. We see that DSEBM-e achieves the highest mean precision and mean F1 score for most cases while DSEBM-r achieves the second best, both beating the competing methods with a margin. HMM shows high precision rates but low recall rates, resulting in a low $F_1$ score. OCCRF is the second best performing method, following our two DSEBM variants, due to its ability to capture the temporal information with CRFs. On FITNESS, DSEBMs improves over 4\%, 6\% and 5\% on mean precision, mean recall and mean $F_1$ over the other baselines. Similar trends can be seen in Figure \ref{tab:sequential:result} (with $\rho$ varying from 0.1 to 0.4). All these results demonstrate DSEBMs' ability to benefit from the rich temporal (with large \rm{Avg(t)}) information, thanks to the underlying RNN. 

\begin{figure*}[t]
\centering
\includegraphics[width=\textwidth,height=0.24\textwidth]{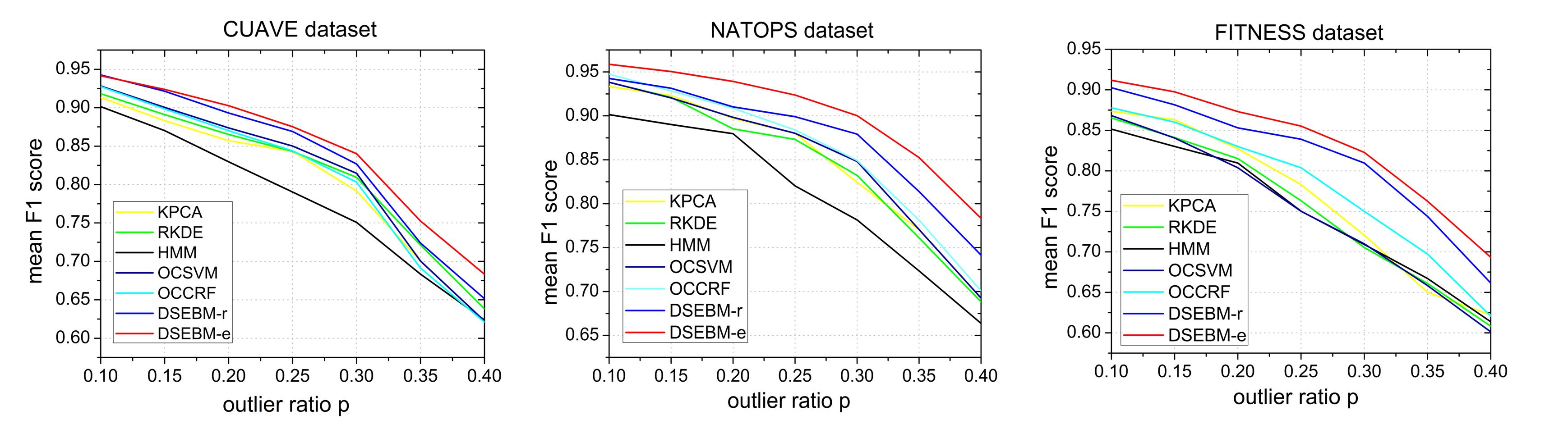}
\caption{The  means $F_1$ scores on the three sequential datasets with outlier ratio from 0.1 to 0.4.}
\label{fig:f1:seq}
\end{figure*}

\begin{table*}[ht]
\small
\centering
\begin{tabular}{|c|ccc|ccc|ccc|}
\hline
\multirow{2}{*}{\textbf{Method}} & \multicolumn{3}{c|}{CUAVE ($\rho=0.3$)}  & \multicolumn{3}{c|}{NATOPS ($\rho=0.3$)} & \multicolumn{3}{c|}{FITNESS ($\rho=0.3$)} \\
\cline{2-10}
& \textbf{mPrec} & \textbf{mRec} & \textbf{m$F_1$} & \textbf{mPrec} & \textbf{mRec} & \textbf{m $F_1$} & \textbf{mPrec} & \textbf{mRec} & \textbf{m$F_1$} \\
\hline
Kernel PCA & 0.8531 & 0.7429 & 0.7916 & 0.8532 & 0.7845 & 0.8244 & 0.7518 & 0.6733 & 0.7206 \\ 
RKDE & 0.8479 & 0.7713 & 0.8094 & 0.8445 & 0.7139 & 0.7822 & 0.7417 & 0.6632 & 0.7054 \\
HMM & \bf 0.8793 & 0.6202 & 0.7507 & 0.8693 & 0.5724 & 0.7215 & 0.7883 & 0.6321 & 0.7092 \\
OC-SVM & 0.8512 & 0.7726 & 0.8147 & 0.8741 & 0.8236 & 0.8481 & 0.7472 & 0.6774 & 0.7099 \\
OCCRF & 0.8329 & 0.7698 & 0.8032 & 0.8795 & 0.8382 & 0.8493 & 0.8088 & 0.7173 & 0.7501\\
DSEBM-r & 0.8692 & 0.7881 & 0.8268 & 0.9035 & 0.8655 & 0.8792 & 0.8425 & 0.7710 & 0.8097 \\
DSEBM-e &  0.8754 & \bf 0.8033 & \bf 0.8402 & \bf 0.9178 & \bf 0.8856 & \bf 0.9022 & \bf 0.8533 & \bf 0.7873 & \bf 0.8228 \\
 \hline
\end{tabular}
\caption{\textbf{CUAVE, NATOPS, FITNESS}: mean precision (mPrec), mean recall (mRec) and mean $F_1$ (m$F_1$) over over the sequential data sets of seven methods. For each column, the best result is shown in boldface.}
\label{tab:sequential:result}
\end{table*}

\begin{figure*}[t]
\centering
\includegraphics[width=\textwidth,height=0.24\textwidth]{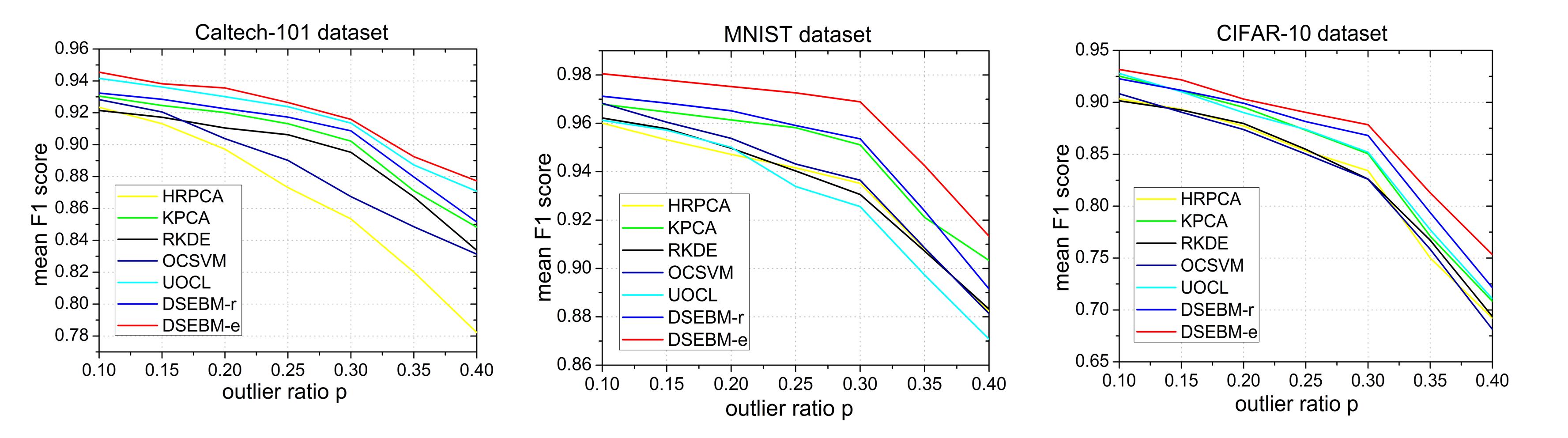}
\caption{The means $F_1$ scores on the three image datasets with outlier ratio from 0.1 to 0.4.}
\label{fig:f1:image}
\end{figure*}

\begin{table*}[!htbp]
\small
\centering
\begin{tabular}{|c|ccc|ccc|ccc|}
\hline
\multirow{2}{*}{\textbf{Method}} & \multicolumn{3}{c|}{Caltech-101 ($\rho=0.3$)} & \multicolumn{3}{c|}{MNIST ($\rho=0.3$)}  & \multicolumn{3}{c|}{CIFAR-10 ($\rho=0.3$)} \\
\cline{2-10}
& \textbf{mPrec} & \textbf{mRec} & \textbf{m$F_1$} & \textbf{mPrec} & \textbf{mRec} & \textbf{m $F_1$} & \textbf{mPrec} & \textbf{mRec} & \textbf{m$F_1$} \\
\hline
HR-PCA & 0.8735 & 0.8025 & 0.8534 & 0.9278 & 0.9493 & 0.9352 & 0.8459 & 0.8217 & 0.8342 \\ 
Kernel PCA & 0.9005 & 0.9091 & 0.9022 & 0.9427 & 0.9576 & 0.9511  & 0.8552 & 0.8473 & 0.8506 \\
RKDE  & 0.8904 & 0.8995 & 0.8952 & 0.9377 & 0.9218 & 0.9306 & 0.8319 & 0.8202 & 0.8261 \\
OC-SVM & 0.8598 & 0.8772 & 0.8674 & 0.9245 & 0.9432 & 0.9356 & 0.8332 & 0.8206 & 0.8259 \\
UOCL & 0.9203 & 0.9076 & 0.9135 & 0.9342 & 0.9198 & 0.9256 & 0.8613 & 0.8442 & 0.8520 \\
DSEBM-r & \bf 0.9184 & 0.9037 & 0.9077 & 0.9597 & 0.9503 & 0.9536 & 0.8742 & 0.8603 & 0.8681 \\
DSEBM-e & 0.9175 & \bf 0.9042 & \bf 0.9159 & \bf 0.9788 & \bf 0.9616 & \bf 0.9689 & \bf 0.8873 & \bf 0.8647 & \bf 0.8784 \\
\hline
\end{tabular}
\caption{\textbf{Caltech-101, MNIST, CIFAR-10} datasets: mean precision (mPrec), mean recall (mRec) and mean $F_1$ (m$F_1$) over over the image data sets of eight methods. For each column, the best result is shown in boldface.}
\label{tab:image:result}
\end{table*}

\subsection{Spatial Data}
We use three public image datasets: Caltech-101, MNIST and CIFAR-10 for this sub task. On Caltech-101, we choose 11 object categories as inliers,  each of which contains at least 100 images, and sample outlier images with a proportion $0.1 \le \rho \le 0.4$ from the other categories. On MNIST and CIFAR-10, we use images from a single category as inliers, and sample images from the other categories with a proportion $0.1 \le \rho \le 0.4$. Each dataset is split into a training and testing set with a ratio of 2:1. We compare DSEBMs (with one convolutional layer + one pooling layer + one fully connected layer) with several baseline methods including: High-dimensional Robust PCA (HR-PCA), Kernel PCA (KPCA), Robust Kernel Density Estimator (RKDE), One-Class SVM (OC-SVM) and Unsupervised One-Class Learning (UOCL) \cite{DBLP:conf/cvpr/LiuHS14}. All the results are shown in Table \ref{tab:image:result} with $\rho=0.3$. We see that DSEBM-e is the best performing method overall in terms of mean recall and mean $F_{1}$, with particularly large margins on large datasets (MNIST and CIFAR-10). Measured by $F_{1}$, DSEBM-e improves 3.5\% and 2.3\% over the best-performing baselines. Figure \ref{fig:f1:image} with $\rho$ varying from 0.1 to 0.4 also demonstrates consistent results. 

\begin{figure*}[!tb]
\centering
\includegraphics[width=0.95\textwidth,height=0.18\textwidth]{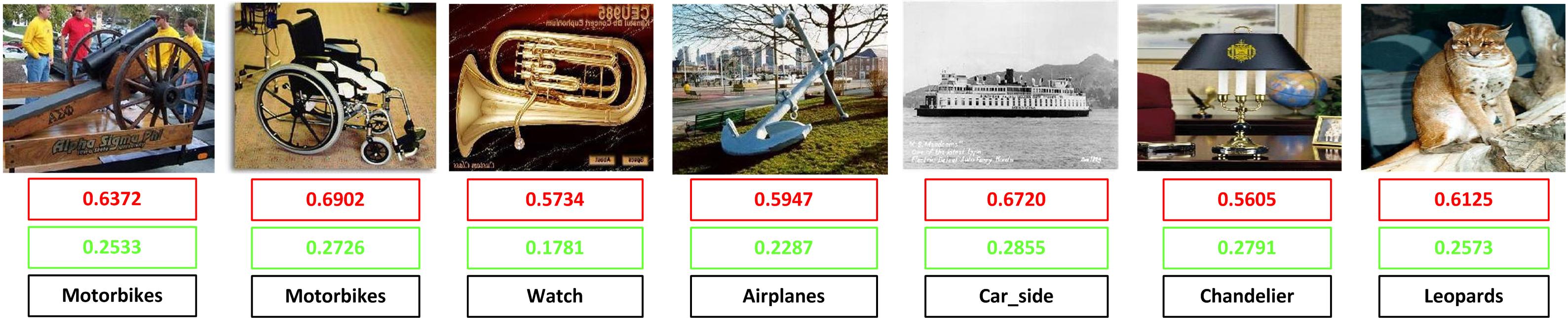}
\caption{Seven outliers from the Caltech-101 dataset. Each column is the image followed by its energy score (displayed in red), reconstruction error (displayed in green) and the inlier class name. The thresholds for DSEBM-e and DSEBM-r are $\rm{E_{th}}=0.4882$ and $\rm{Error_{th}}=0.5125$, respectively. Samples with $\rm{E(x) > E_{th}}$ and $\rm{Error(x) > Error_{th}}$ are regarded as outliers by DSEBM-e and DSEBM-r, respectively.}
\label{fig:wearable_example}
\end{figure*}

\subsection{Energy VS. Reconstruction Error}
In terms of the two decision criteria of DSEBM, we observe that DSEBM-e consistently outperforms DSEBM-r on all the benchmarks except for the Thyroid dataset. This verifies our conjecture that the energy score is a more accurate decision criterion than reconstruction error. In addition, to gain further insight on the behavior of the two criteria, we demonstrate seven outliers selected from the Caltech-101 benchmark in Figure \ref{fig:wearable_example}. For each image, the energy scores are displayed at the second row in red, followed by the reconstruction error displayed in green and the correct inlier class. Interestingly, all the seven outliers are visually similar to the inlier class and have small reconstruction errors (compared with the threshold). However, we are able to successfully identify all of them with energy (which are higher than the energy threshold). 

\section{Related Work}
There has been a large body of work concentrating on anomaly detection \cite{survey}, noticeably: (1) the reconstruction based methods such as PCA and Kernel PCA, Robust PCA and Robust Kernel PCA; (2) the probability density based methods, including parametric estimators \cite{Zimek:2012:SUO:2388912.2388917} and nonparametric estimators such as the kernel density estimator (KDE) and the more recent robust kernel density estimator (RKDE); (3) methods of learning a compact data model such that as many as possible normal samples are enclosed inside, for example, one-class SVM and SVDD. Graham \textit{et al.} proposed a method based on autoencoder \cite{conf/icdm/WilliamsBHHG02}. However, all the methods above are static in nature which does not assume the structure of data. Two types of data are extensively studied in sequential anomaly detection: sequential time series data and event data. Sun \textit{et al.} proposes a technique that uses Probabilistic Suffix Trees (PST) to find the nearest neighbors for a given sequence to detect sequential anomalies in protein sequences \cite{citeulike:1129276}. Song \textit{et al.} presents a one class conditional random fields method for general sequential anomaly detection tasks \cite{Song:2013:OCR:2540128.2540370}. Our model is significantly different from the above mentioned methods, where our use of RNN encoded EBM gives us much modeling power and statistical soundness at the same time. Among the few approaches designed for spatial data, \cite{DBLP:journals/ieicet/NamS15} proposes to use CNNs in least-squares direct density-ratio estimation, and demonstrated its usefulness in inlier-based outlier detection of images. Despite the usage similar use of CNNs, our work takes a very different path by directly modeling the density.
Methodology-wise, there is also a recent surge of training EBMs with score matching \cite{smae1,smae2, kingma2010}. However, most of them are constrained to shallow models, thus limiting their application to relatively simple tasks.

\section{Conclusion}
We proposed training deep structured energy based models for the anomaly detection problem and extended EBMs to deep architectures with three types of structures: fully connected, recurrent and convolutional. To significantly simplify the training procedure, score matching is proposed in stead of MLE as the training algorithm. In addition, we have investigated the proper usage of DSEBMs for the purpose of anomaly detection, in particular focusing on two decision criteria: energy score and reconstruction error. Systematic experiments are conducted on three types of datasets: static, sequential and spatial, demonstrating that DSEBMs consistently match or outperform the state-of-the-art anomaly detection algorithms. To be best of our knowledge, this is the first work that extensively evaluates deep structured models to the anomaly detection problem.

\bibliography{example_paper}
\bibliographystyle{icml2016}

\end{document}